# Monaural Multi-Talker Speech Recognition using Factorial Speech Processing Models

Mahdi Khademian [a], Mohammad Mehdi Homayounpour [a, *]

[a] Laboratory for Intelligent Multimedia Processing (LIMP), Computer Engineering and IT Department, Amirkabir University of Technology, Tehran, Islamic Republic of Iran

**Abstract**

A Pascal challenge entitled monaural multi-talker speech recognition was developed, targeting the problem of robust automatic speech recognition against speech like noises which significantly degrades the performance of automatic speech recognition systems. In this challenge, two competing speakers say a simple command simultaneously and the objective is to recognize speech of the target speaker. Surprisingly during the challenge, a team from IBM research, could achieve a performance better than human listeners on this task. The proposed method of the IBM team, consist of an intermediate speech separation and then a single-talker speech recognition. This paper reconsiders the task of this challenge based on gain adapted factorial speech processing models. It develops a joint-token passing algorithm for direct utterance decoding of both target and masker speakers, simultaneously. Comparing it to the challenge winner, it uses maximum uncertainty during the decoding which cannot be used in the past two-phased method. It provides detailed derivation of inference on these models based on general inference procedures of probabilistic graphical models. As another improvement, it uses deep neural networks for joint-speaker identification and gain estimation which makes these two steps easier than before producing competitive results for these steps. The proposed method of this work outperforms past super-human results and even the results were achieved recently by Microsoft research, using deep neural networks. It achieved 5.5% absolute task performance improvement compared to the first super-human system and 2.7% absolute task performance improvement compared to its recent competitor.

* Corresponding author: Mohammad Mehdi Homayounpour
 Email: homayoun@aut.ac.ir
 Tel: +98 21 64542722

# 1. Introduction

Robustness of automatic speech recognition systems (ASR) against diverse speech processing environments and adverse disturbing noises still remains as one of the important research areas in speech recognition systems (Baker et al., 2009; Li et al., 2014). Among all diversities and conditions in everyday environments which ASR systems must manage, dealing with the Babble noise and presence of competing speakers is one of the challenging problems of these systems. This problem is known as the cocktail party problem (Haykin and Chen, 2005) in which a person (or a system) wants to focus and follow the conversation of a speaker in a place where some people talk simultaneously. Roughly speaking, two groups of approaches are developed for addressing this problem. Approaches in the first group incorporate signals captured from several microphones or capturing channels and perform low level signal processing techniques such as the beam forming and blind source separation which reduce footprints of the competing audio sources. These approaches accomplish their speech processing tasks using the improved captured and processed signals. Approaches in the second group use only one recording channel and perform high level speech processing and machine learning techniques and accomplish their tasks in the presence of the competing audio sources which seems to be more challenging.

An interesting competition entitled "Monaural speech separation and recognition challenge", addressing the challenges related to the second group of approaches was developed in 2006 (Cooke et al., 2010). In this challenge, two competing speakers are simultaneously issuing a command and the objective is to recognize the command of the target speaker. The task uses a simple grammar for commands and it has a small vocabulary.

The problem of monaural multi-talker speech recognition becomes more difficult when speech of the masker speaker has higher energy than the target. The problem becomes worse when the masker speaker voice is similar to the target voice; i.e. when the two speakers have the same gender or two speakers are the same. Several teams attended this competition with different techniques for handling the problem (Cooke et al., 2010). Among the competitors, surprisingly, a team from IBM research presented a technique that outperforms the other techniques and even human listeners (Hershey et al., 2010). In their work, the task was accomplished in three main steps. First, the identity of speakers and gain is estimated using high resolution Gaussian Mixture Models (GMM) as speaker models. In the main step, speech of both speakers are separated from the mixed-speech signal using factorial speech processing models. In this step, expected value of source features given the observed feature and joint acoustic states are considered for source estimation, then source separation is performed. In the third step, two separated speeches are decoded using a single-talker recognition system (Hershey et al., 2010). This team further developed their system to support mixture of speeches of more than two speakers and separate their voices only by one recording channel! Later researches continue to work on this dataset. To the best of our knowledge, only one work outperforms IBM's super-human results from the Microsoft research. This work incorporates a pair of Deep Neural Networks for acoustic inference over semi-joint hidden Markov models (HMM) (Weng et al., 2015); a network for the generation of senone posteriors of high energy utterances (or instantaneous high energy frames) and a network for low energy utterances. Then they perform decoding over these senone posteriors.

The method presented in this paper is a model based approach based on factorial speech processing models for recognizing monaural mixed-speech signals which is applied for the "Monaural speech separation and recognition challenge". It directly performs joint-decoding over the model to decode both utterances of the target and masker speaker, simultaneously. Direct speech recognition of this work is done by a joint-decoder which is developed by extending the token passing algorithm to support inference over factorial speech processing models constructed with grammar and dictionary. While joint-decoding over the models with grammar and dictionary increases complexity of decoder, at the same time it provides significant performance improvement over the past developed systems. Additionally, this work uses Deep Neural Networks for speaker identification and gain estimation as one important step for determining audio sources of the factorial model. Moreover, this paper presents a detailed inference procedure over the factorial models using the general inference procedures of probabilistic graphical models.

After this introduction, the next section briefly describes the challenge. It also presents a detailed description of steps for applying factorial speech processing models to this challenge, first by deriving inference procedures for the factorial speech processing models, then developing the joint-token passing algorithm for performing the joint-decoding. Section 3 describes the methods for determining and adapting source models. Section 4 presents experiments, scoring procedure and results. Section 5 will conclude the paper.

## 2. Factorial speech processing models for single channel speech recognition

The objective of the monaural speech separation and recognition challenge is to recognize some keywords of a target speaker from a mixed-speech of the target and a masker speaker (Cooke et al., 2010). Mixed-speech signals of this task are artificially created from speech materials of the Grid corpus (Cooke et al., 2006). This corpus contains simple six-word slot commands from 34 different speakers. Each command is a sequence of command words, color, preposition, a letter, a digit and an adverb depicted in Fig. 1.

($command_word  $color  $preposition **$letter $digit** $adverb)

$command_word = bin | lay | place | set ;
$color = white | blue | green | red ;
$preposition = at | by | in | with ;
**$letter** = a | b | c | d | e | f | g | h | i | j | k | l | m | n | o | p | q | r | s | t | u | v | x | y | z ;
**$digit** = one | two | three | four | five | six | seven | eight | nine | zero ;
$adverb = again | now | please | soon ;

**Fig. 1. Task grammar consists of a sequence of six-word slots (w is not included in letters). Grammar is provided in the HTK HParse (Young et al., 2009) grammar definition syntax. The recognized letter and digit of the target speaker are used for scoring.**

Mixed-speech signals are created by selecting two utterances from the Grid corpus, one for the target speaker and the other for the masker. The target speaker always uses "white" as the command color and the masker does not. This is the clue for discrimination of the target and masker voices. Two speech signals are mixed by the following time domain relation:

$$y = x^a + gx^b \qquad (1)$$

in which $x^a$ is the speech signal for the target and $x^b$ is for the masker. The challenge is designed for different signal energy ratios of the target and masker which is called Target to Masker Ratio (TMR). This is adjusted by the gain coefficient, $g$, in (1). Objective keywords of the task are the letter and digit of the target speaker which is used for scoring. The next sub-section provides an overview of factorial models of speech processing, which are used in this work for performing the task. Then the inference and decoding procedures of these models are presented.

## 2.1. Factorial speech processing models

Factorial models of speech processing are generative models for modeling the combination of multiple audio sources into one (or even multiple) observable mixed-audio signals. It is applicable for robust-ASR systems (Hershey et al., 2012) and is tailored exactly for the challenge of this paper. Fig. 2 shows the graphical model of a factorial speech processing model. This model is based on factorial hidden Markov models which are used for modeling processes with multiple independent underlying Markov chains (Koller and Friedman, 2009). In this figure, the two source Markov processes are the speech process of speaker $a$ and $b$. Conditional probability distribution (CPD) of the Markov chain of the audio sources are $p(s_{t+1}^a|s_t^a)$ and $p(s_{t+1}^b|s_t^b)$ which are modeled parametrically by stochastic matrices. Each chain of a factorial speech processing model contains an HMM for modeling its audio source which is known as acoustic modeling (Young, 1996) in conventional speech recognition applications. These are shown in Fig. 2, by the dashed boxes around each audio source (source $a$ and $b$).

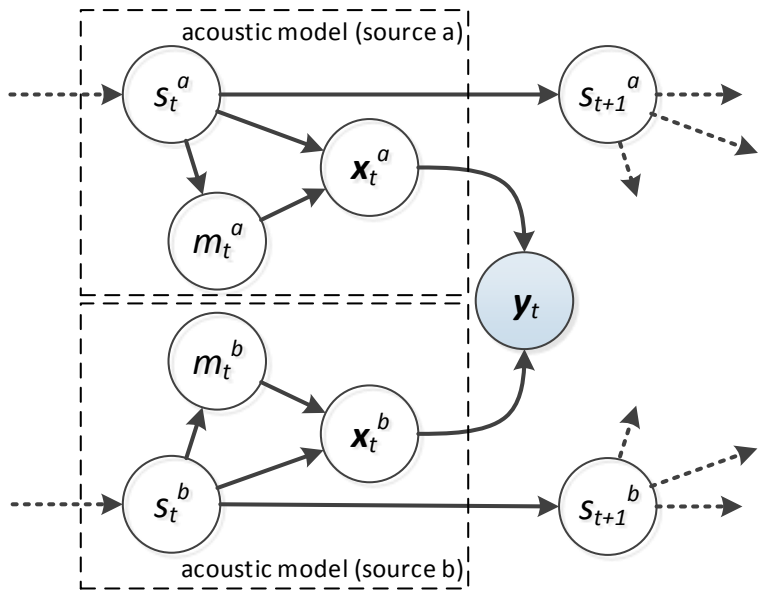

**Fig. 2.** Factorial Speech Processing Model for recognition of two audio sources only by one recording channel. In this graphical model, $y_t$ is the feature vector of captured signal at time frame $t$ and is observable (other variables are hidden), $s_t^a$ and $s_t^b$ are hidden states of the two audio sources, $m_t^a$ and $m_t^b$ are mixture components of the audio sources when the Gaussian mixture model is used for acoustic modeling and $x_t^a$ and $x_t^b$ are hidden feature vectors of the audio sources.

Conventionally, in speech processing applications, Gaussian mixture models are used for observation distributions of HMM and left-to-right topology is used for modeling the Markov chain of that HMM. The observation probability distribution of HMMs is modeled by the following CPD:

$$p(x_t^a|s_t^a) = \sum_{m_t^a} p(x_t^a|m_t^a, s_t^a) \, p(m_t^a|s_t^a) \tag{2}$$

where $p(x_t^a|m_t^a = i_m, s_t^a = i_s) = \mathcal{N}(x_t^a; \boldsymbol{\mu}_{(i_s,i_m)}, \boldsymbol{\Sigma}_{(i_s,i_m)})$ is the $i_m$ Gaussian component of the GMM of state $i_s$, and $p(m_t^a|s_t^a)$ models the component weights by a stochastic matrix. The observation model of the second chain is similar to the first one.

Factorial speech processing models have additional CPD for combining source features comparing them to factorial hidden Markov models which is called the acoustic interaction function. This CPD provides a probabilistic relationship between the source and the combined features, i.e. $p(y_t|x_t^a, x_t^b)$. In these models, source features are not observable but the combined feature is. By the acoustic interaction function we can infer the posterior distribution of source features and then the posterior distribution of source states in an "evidential reasoning" pattern (Koller and Friedman, 2009). The number of Markov chains in factorial models of speech processing depends on the number of audio sources in the mixed-signal.

For the recognition task of monaural speech separation and recognition challenge, we can consider one audio source of the factorial model as the clean source model of the target speaker and the other as the model of masker. Then based on mixed-signal feature vectors, we can infer acoustic states of the audio sources which are used for decoding. The next sub-section describes the way that inference is done over the factorial models of speech processing.

## 2.2. Inference

The objective of inference in models in the form of Fig. 2 is to find the most probable states of the sources in a period of time given the observed feature vectors of that period:

$$s_{1:T}^{*a,b} = \underset{s_{1:T}^{a,b}}{\operatorname{argmax}} \, p(s_{1:T}^{a,b}|y_{1:T}) \tag{3}$$

But the main objective of inference in our task is to find the sequence of spoken words of each speaker; more specifically, the mentioned letter and digit of the target speaker. Decoding the most probable acoustic states into the sequence of spoken words is done by a joint-decoder. In this sub-section, acoustic and temporal inference over the factorial model is described and in the next sub-section the decoding procedure is discussed.

### 2.2.1. Acoustic inference

The graphical model of Fig. 2 is not used directly in the applications. In fact, an intermediate step during the inference is marginalizing-out hidden feature vectors of audio sources ($x^a$ and $x^b$). This step will merge three CPDs including $p(x^a|s^a, m^a)$, $p(x^b|s^b, m^b)$ and $p(y|x^a, x^b)$ into state conditional likelihood, $p(y|s^a, m^a, s^b, m^b)$. Probabilistically we have:

$$p(y|s^a, m^a, s^b, m^b) = \iint p(y|x^a, x^b)p(x^a|s^a, m^a)p(x^b|s^b, m^b)dx^a dx^b \qquad (4)$$

As a result, the model of Fig. 2 will be simplified to the graphical model of Fig. 3. This form of marginalization is seen before in noise-robust automatic speech recognition when one audio source is clean speech and the other is the disturbing noise (Hershey et al., 2012). Depending on the feature space and source models, several approaches are suggested for the calculation of (4). For example, for high resolution power spectral features, the max-model (Roweis, 2003) is used for marginalizing-out source feature vectors. This feature space is usually used for enhancement applications. The objective of our task is to recognize utterances of the audio sources (speakers) where the audio sources are modeled by GMMs. Usually for this application, MFCC features are used for source modeling. By using an appropriate mismatch function (Gales and Young, 1996) which combines source features into the observed feature vector, we can approximate each state conditional distribution by one Gaussian. This technique is known as approximation by the vector Taylor series (Moreno et al., 1996), VTS, which will be discussed later.

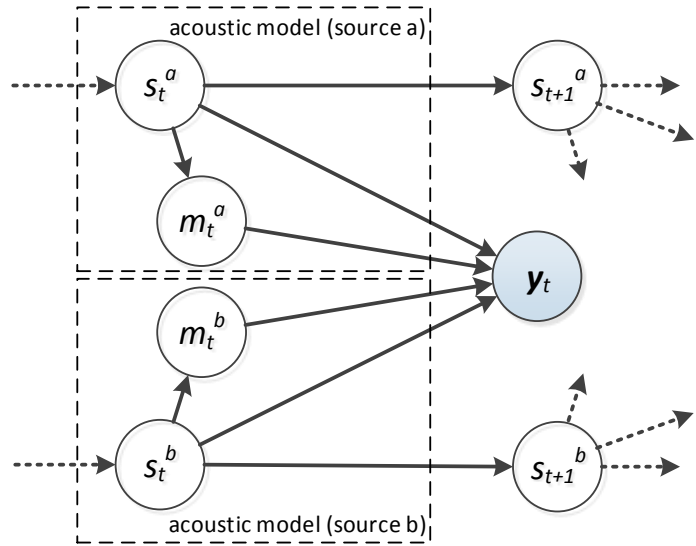

**Fig. 3.** Simplified Factorial Hidden Markov Model of Fig. 2 by marginalizing-out source feature vectors

### 2.2.2. VTS based acoustic inference

Returning to the time domain expression (1) for creating the mixed-speech signal, here the expressions for combining clean source features and then state-conditional likelihoods of mixed-speech features are developed. The gain coefficient is omitted temporarily and will be reconsidered in section 3.2, in details. After framing, windowing and by use of short-term discrete Fourier transform we have the following relation between the frames' complex feature vectors:

$$\boldsymbol{y}_t = \boldsymbol{x}_t^a + \boldsymbol{x}_t^b \tag{5}$$

where vector elements contain different frequency bins. In the power spectrum we have:

$$|\boldsymbol{y}_t|^2 = |\boldsymbol{x}_t^a|^2 + |\boldsymbol{x}_t^b|^2 + 2|\boldsymbol{x}_t^a||\boldsymbol{x}_t^b|\cos\boldsymbol{\theta} \tag{6}$$

in which $\boldsymbol{\theta}$ is the phase difference between $\boldsymbol{x}_t^a$ and $\boldsymbol{x}_t^b$ complex vectors at different frequency bins ($t$ subscript is removed in subsequent expression and provided as needed for notational brevity). In the terminology of noise-robust ASR, this expression is called "mismatch function" when one of the source signals is the disturbing noise (Gales and Young, 1996). Then, by using Mel scale averaging filters we have:

$$\widetilde{\boldsymbol{y}} = \mathbf{W}|\boldsymbol{y}|^2, \widetilde{\boldsymbol{x}}^a = \mathbf{W}|\boldsymbol{x}^a|^2, \widetilde{\boldsymbol{x}}^b = \mathbf{W}|\boldsymbol{x}^b|^2$$

in which $\mathbf{W}$ is the weighing matrix for Mel filters (each row contains weighting elements for each Mel filter). Now for the Mel filter bank features, equation (6) is transformed to:

$$\widetilde{\boldsymbol{y}} = \widetilde{\boldsymbol{x}}^a + \widetilde{\boldsymbol{x}}^b + 2\boldsymbol{\alpha}\sqrt{\widetilde{\boldsymbol{x}}^a \widetilde{\boldsymbol{x}}^b} \tag{7}$$

where $\boldsymbol{\alpha}$ is called the phase factor and is equal to:

$$\boldsymbol{\alpha} = \frac{\mathbf{W}|\boldsymbol{x}^a||\boldsymbol{x}^b|\cos\boldsymbol{\theta}}{\sqrt{\widetilde{\boldsymbol{x}}^a \widetilde{\boldsymbol{x}}^b}} \tag{8}$$

Considering random values for $\boldsymbol{\theta}$ in $[-\pi, \pi]$, alpha becomes stochastic in the range of $[-1,1]$ (Van Dalen, 2011). Moreover $\boldsymbol{\alpha}$ can be considered independent from source feature vectors by an additional simplifying assumption (Leutnant and Haeb-Umbach, 2009). Thus, it can be considered as an internal independent stochastic variable in the mismatch function. It is traditionally ignored in many applications (Van Dalen, 2011) or considered as a constant for all frequency bins (Li et al., 2009). By taking the logarithm and left multiplying the [truncated] DCT matrix, we have the following relation between the feature vectors in the Cepstrum domain:

$$^c\boldsymbol{y} = \mathbf{C}\log\left(\exp(\mathbf{C}^{-1}{}^c\boldsymbol{x}^a) + \exp(\mathbf{C}^{-1}{}^c\boldsymbol{x}^b) + 2\alpha\exp\left(\frac{1}{2}\mathbf{C}^{-1}({}^c\boldsymbol{x}^a + {}^c\boldsymbol{x}^b)\right)\right) \quad (9)$$

where $^c$ in $^c\boldsymbol{y}$ denotes Cepstral features ($^c$ will be removed in subsequent expressions, from now on, all feature vectors are considered as MFCC features). Equation (9) is known as the most applicable mismatch function in noise-robust ASR where one of the audio sources is considered as the disturbing noise; i.e. $\boldsymbol{y} = \mathbf{f}(\boldsymbol{x}^a, \boldsymbol{x}^b, \alpha)$. This equation constructs a non-linear relationship between feature vectors of the audio sources and observable features. It can be considered as a deterministic CPD of $p(\boldsymbol{y}\,|\boldsymbol{x}^a, \boldsymbol{x}^b)$ in Fig. 2. A linear approximation of (9) could be established by first-order vector Taylor series expansion around $\boldsymbol{x}_0^a$, $\boldsymbol{x}_0^b$ and considering a constant alpha as:

$$\boldsymbol{y} = \mathbf{f}(\boldsymbol{x}^a, \boldsymbol{x}^b, \alpha) \cong \boldsymbol{f}_0 + \mathbf{G}(\boldsymbol{x}^a - \boldsymbol{x}_0^a) + \mathbf{H}(\boldsymbol{x}^b - \boldsymbol{x}_0^b) \quad (10)$$

in which $\boldsymbol{f}_0 = \boldsymbol{f}(\boldsymbol{x}_0^a, \boldsymbol{x}_0^b, \alpha)$, $\mathbf{G} = \left.\frac{\partial f}{\partial x^a}\right|_{x_0^a, x_0^b, \alpha}$ and $\mathbf{H} = \left.\frac{\partial f}{\partial x^b}\right|_{x_0^a, x_0^b, \alpha}$.

Expression (10) can be viewed as the sum of two transformed vector random variables. By assuming that $\boldsymbol{y}$ has Gaussian distribution and considering the expansion point as the source variable means, the Gaussian parameters become:

$$\boldsymbol{\mu}_y = \boldsymbol{f}(\boldsymbol{\mu}_{x^a}, \boldsymbol{\mu}_{x^b}, \alpha) \quad (11)$$

$$\boldsymbol{\Sigma}_y = \mathbf{G}\boldsymbol{\mu}_{x^a}\mathbf{G}^T + \mathbf{H}\boldsymbol{\mu}_{x^b}\mathbf{H}^T \quad (12)$$

At this point, only inference over the static part of feature vectors is done. For performing inference over the dynamic coefficients, consider the following expression for discrete differentiation of feature vectors:

$$\Delta \boldsymbol{y}_t = \sum_i w_i \boldsymbol{y}_{t-i}, \sum w_i = 0 \quad (13)$$

in which $w_i$s are weights for differentiation. For example, for a context window of size 5, $w_i$s are $\{-1/4, -1/2, 0, 1/2, 1/4\}$. Considering expression (13), assuming that dynamic coefficients can also be modeled by a Gaussian and by one practical approximation, Gaussian parameters for delta and acceleration coefficients can be extracted similar to static parts as follows:

$$\boldsymbol{\mu}_{\Delta y} = \mathbf{G}\boldsymbol{\mu}_{\Delta x^a} + \mathbf{H}\boldsymbol{\mu}_{\Delta x^b} \quad (14)$$

$$\boldsymbol{\Sigma}_{\Delta y} = \mathbf{G}\boldsymbol{\mu}_{\Delta x^a}\mathbf{G}^T + \mathbf{H}\boldsymbol{\mu}_{\Delta x^b}\mathbf{H}^T \quad (15)$$

$$\boldsymbol{\mu}_{\Delta\Delta y} = \mathbf{G}\boldsymbol{\mu}_{\Delta\Delta x^a} + \mathbf{H}\boldsymbol{\mu}_{\Delta\Delta x^b} \quad (16)$$

$$\boldsymbol{\Sigma}_{\Delta\Delta y} = \mathbf{G}\boldsymbol{\mu}_{\Delta\Delta x^a}\mathbf{G}^T + \mathbf{H}\boldsymbol{\mu}_{\Delta\Delta x^b}\mathbf{H}^T \quad (17)$$

For a detailed derivation, the reader is referred to the appendix of (Li et al., 2009). Now the acoustic inference can be done by the following expression:

$$p(\mathbf{y}|s^a = i_s, m^a = i_m, s^b = j_s, m^b = j_m) = \mathcal{N}\big(\mathbf{y}_s; \boldsymbol{\mu}_{sy(i_s,i_m,j_s,j_m)}, \boldsymbol{\Sigma}_{sy(i_s,i_m,j_s,j_m)}\big) \times$$
$$\mathcal{N}\big(\mathbf{y}_\Delta; \boldsymbol{\mu}_{\Delta y(i_s,i_m,j_s,j_m)}, \boldsymbol{\Sigma}_{\Delta y(i_s,i_m,j_s,j_m)}\big) \times \mathcal{N}\big(\mathbf{y}_{\Delta\Delta}; \boldsymbol{\mu}_{\Delta\Delta y(i_s,i_m,j_s,j_m)}, \boldsymbol{\Sigma}_{\Delta\Delta y(i_s,i_m,j_s,j_m)}\big) \quad (18)$$

in which $\mathbf{y} = [\mathbf{y}_s, \mathbf{y}_\Delta, \mathbf{y}_{\Delta\Delta}]'$. The expansion point for calculating means and covariance matrices are determined by the source states and GMM components. For example, $\boldsymbol{\mu}_{\Delta y(i_s,i_m,j_s,j_m)} = \mathbf{G}\boldsymbol{\mu}_{\Delta y_{i_s,i_m}}\mathbf{G}^T + \mathbf{H}\boldsymbol{\mu}_{\Delta y_{j_s,j_m}}\mathbf{H}^T$. Now, state conditional likelihoods for different joint-source states and GMM components can be calculated by (18). At this moment, we are prepared to perform temporal inference in the graphical model of Fig. 3 which is covered in the next sub-section.

### 2.2.3. Temporal inference

For extracting exact temporal inference expressions, at first, a clique tree is constructed from the simplified factorial model of Fig. 3. This clique tree is depicted in Fig. 4 which is an arbitrary tree constructed by first eliminating the state variable of source b and then a.

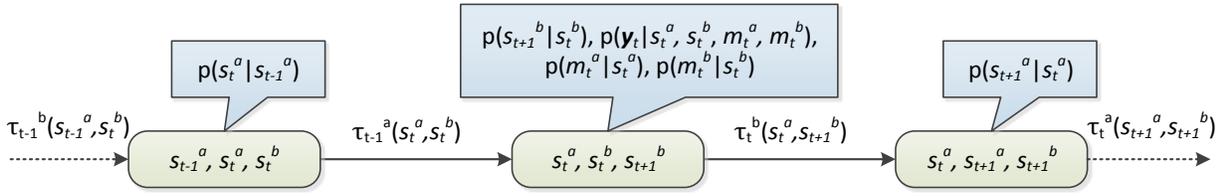

**Fig. 4.** A clique tree constructed from the simplified factorial hidden Markov model of Fig. 3. Involved factors in each inference step are listed in the callout boxes. Involved variables in each inference step are listed in the boxes. Eliminated variables by the max operator are relative compliment of involved variables in all variables in the involved factors. Newly generated factors are noted over the arrows.

Based on the constructed clique tree, the following recursions are extracted to do temporal inference:

$$\tau_t^b(s_t^a, s_{t+1}^b) = \max_{s_t^b, m_t^a, m_t^b} \tau_{t-1}^a(s_t^a, s_t^b) p(s_{t+1}^b|s_t^b) p(m_t^a|s_t^a) p(m_t^b|s_t^b) p(\mathbf{y}_t|s_t^a, m_t^a, s_t^b, m_t^b) \quad (19)$$

$$\tau_t^a(s_{t+1}^a, s_{t+1}^b) = \max_{s_t^a} \tau_t^b(s_t^a, s_{t+1}^b) p(s_{t+1}^a|s_t^a) \quad (20)$$

where in these two recursions, $p(s_{t+1}^a|s_t^a)$ and $p(s_{t+1}^b|s_t^b)$ are state transition matrices of two underlying Markov chains, $p(m_t^a|s_t^a)$ and $p(m_t^b|s_t^b)$ are component weights of GMM observation models of source

HMMs and $p(\mathbf{y}_t|s_t^a, m_t^a, s_t^b, m_t^b)$ is the result of acoustic inference. The importance of breaking this joint-state maximization problem into two single-variable problems is the reduction of operations by a factor of number of chain states. In other words, instead of jointly doing maximization over $\langle s_t^b, s_t^a \rangle$, at each step one source state variable is optimized in a dynamic programming manner (the presence of $m_t^a, m_t^b$ is not important in the first recursion since the number of GMM components are not significant relative to the number of states). The initial factor $\tau_0^a(s_0^a, s_0^b)$ is defined as:

$$\tau_0^a(s_1^a, s_1^b) = p(s_1^a)p(s_1^b) \tag{21}$$

in which $p(s_1^a)$ and $p(s_1^b)$ are state priors of two underlying Markov chains. At each step, a back pointer to the previous maximized state is used for the final backtracking:

$$\psi_t^b(s_t^a, s_{t+1}^b) = \underset{s_t^b}{\mathrm{argmax}} \left( \tau_{t-1}^a(s_t^a, s_t^b) p(s_{t+1}^b|s_t^b) \max_{m_t^a, m_t^b} p(m_t^a|s_t^a) p(m_t^b|s_t^b) p(\mathbf{y}_t|s_t^a, m_t^a, s_t^b, m_t^b) \right) \tag{22}$$

$$\psi_t^a(s_{t+1}^a, s_{t+1}^b) = \underset{s_t^a}{\mathrm{argmax}}\, \tau_t^b(s_t^a, s_{t+1}^b) p(s_{t+1}^a|s_t^a) \tag{23}$$

This algorithm which is derived here using general inference procedures of probabilistic graphical models is called the two-dimensional Viterbi algorithm (Hershey et al., 2010). For more details about clique tree construction and general inference over the graphical models, the reader can refer to (Koller and Friedman, 2009; Murphy, 2002).

### 2.3. Joint-decoding

For the recognition task of the challenge, different decoding methods can be used. In the simplest method, for speech recognition in a single chain, whole word acoustic models can be used for the creation of a composite HMM. This can be done since the challenge has a small vocabulary. The composite HMM is created by concatenation of states of the whole word HMMs, allowing transitions between the different words in the adjacent word-slots. Single chain composite HMM models can now be considered in a factorial model with multiple-chains as source models for performing inference. This method of inference has some limitations. First, this method cannot be used in the tasks with medium and large vocabulary size since considering whole word acoustic models is not applicable in these scenarios. Second, explicit construction of the composite HMM in the tasks with complex grammar and language-models is very difficult. Third, acoustic states in whole word HMMs have overlap with each other which makes acoustic inference on factorial models, inefficient.

The more efficient method is to use sub-word acoustic units. The past proposed method for decoding single chain HMMs with sub-word acoustic units is called token passing algorithm. It is a conceptual framework for decoding in large vocabulary continuous speech recognition tasks (Young, 1996; Young et al., 2009).

In the token passing algorithm, a word lattice network based on the task grammar is created which models allowable transitions between words of the task. Then a lexicon is used for phonetic transcription of words appearing in the network into a sequence of phonemes. Now based on the network and lexicon, a set of hypotheses of phoneme sequences is used for decoding. These hypotheses and their scores are represented by tokens. Each token preserves a history of states using a pointer to its previous token and it has a score which is the acoustic score of the whole state sequence; the states of different hypotheses of phoneme sequences. At the final time-frame, the best sequence is selected for backtracking by considering its score. Therefore, for each time-frame, the acoustic state of phoneme in the phoneme sequence, the phoneme and the active word are determined and feature vectors of the command are decoded into the command words. However, this method is applicable for single chain models.

The token passing framework can be considered for performing the decoding in the factorial models of speech processing, which is called joint-decoding. For the joint-decoding, a new notion of state is constructed to perform the inference. In a single chain phoneme based decoding, word_id in the word lattice network, phoneme_id in the phonetic transcription of the word and acoustic state of phoneme, construct a rich state which can be used in decoding. In multiple chain models, the Cartesian product of this rich state can be considered as a joint-state. In Fig. 5 we can see joint-states of joint-tokens which are represented by the split ovals. Each part of this oval represents a token with a rich state where the token's corresponding word and phoneme are written in the oval instead of their ids. Additionally, phoneme state is denoted in the parenthesis in front of the phoneme name.

The joint-state of each token is used for doing the acoustic and temporal inference. The joint-state conditional likelihood of each frame is calculated using (18). This likelihood is calculated by considering the relation (18) conditioned on the joint-state of the joint-tokens in the active token list. Then the joint-tokens are propagated through the word lattice network, within a word through the phonemes and states of a phoneme to update the list and moving to the next frame. These steps are depicted by their corresponding examples in Fig. 5. The joint-token passing algorithm for phoneme based joint-decoding is simply stated in Fig. 6 to provide a clear insight into the problem.

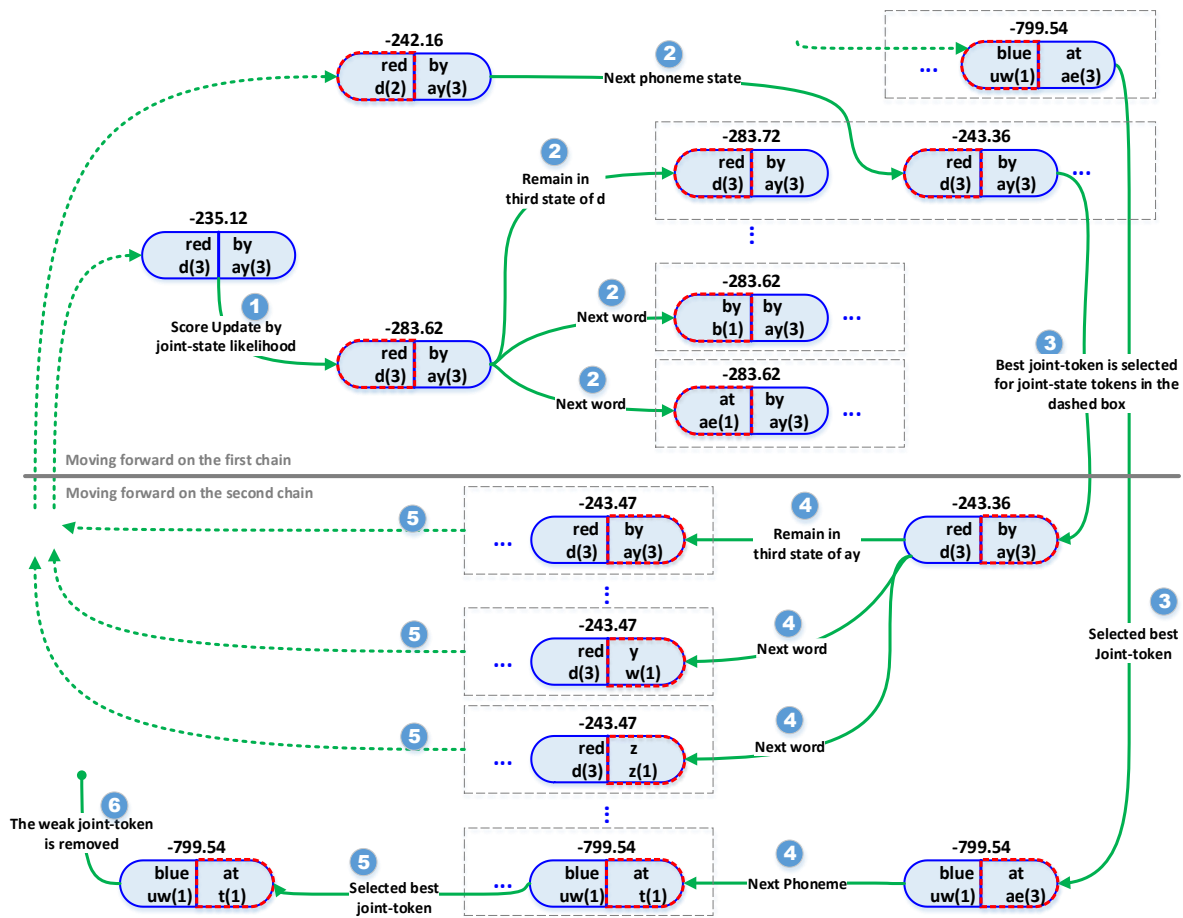

**Fig. 5.** Schematic representation of joint-token passing algorithm. Joint-tokens are represented by split ovals which are copied, updated and discarded in the steps of algorithm to perform the joint-decoding. The active chain on the factorial model designated by dashed part of the ovals. Dummy token scores are stated just above the ovals. Moreover, the related algorithm steps are denoted on the arrows.

As we can see, steps 1-3 in the algorithm are related to the first chain expansion in (19) and steps 4-5 are related to the second chain expansion in (20). The best joint-tokens in each joint-state are selected according to their scores. This is equivalent to the maximization steps of (19) and (20) which is depicted also in Fig. 5 by steps 3 and 5. In this figure, we can see the joint-token <red d(3), by ay(3) -243.36> is selected due to its score which is greater than <red d(3), by ay(3) -283.72>. Moreover after each iteration of algorithm the weak tokens, the tokens with low scores, can be removed. This can be done by selecting some threshold or maintaining active token list size by some limit.

> **Initialization:**
> 1. Extract features from the mixed-speech utterance after framing and windowing.
> 2. Construct the word lattice network of each chain.
> 3. Select the first word in each network. Based on the phonetic transcription of the selected word provided by the lexicon, a token is initialized with the score 0 and a null pointer to its previous token. The token joint-state consists of chain1 and chain2 rich states <(word_id, phoneme_id, phoneme_state), (word_id, phoneme_id, phoneme_state)>. In these rich states, the first phoneme state is the entering state of the hidden Markov model of its corresponding phoneme.
>
> **Main Loop (for each time-frame):**
> 1. Update the token scores by (18) using features of the current time-frame.
> 2. Go forward on the chain 1. This includes creating a copy of the token which may pass over state transitions within a phoneme, going to the next phoneme of a word or going to the next word in the word lattice network. Only rich states related to chain 1 are updated while the pointer to the previous token is preserved. Updating token score should be done by considering the state transition probability within a phoneme (other decoding considerations such as word insertion penalties and language model scores can be considered here for updating token score which are not related to the challenge of this paper).
> 3. Best token in each joint-state is selected and the others are discarded.
> 4. Go forward on the chain 2.
> 5. Best token in each joint-state is selected and the others are discarded.
> 6. Remove the weak tokens (the tokens with low scores).
>
> **Termination:**
> 1. Update the token scores by (18) using features of the last time-frame.
> 2. Find the best token.
> 3. Do backtracking by the back-pointer of the best token until reaching the initial token (the token with a null back-pointer).

**Fig. 6. Phoneme based joint-decoding by joint-token passing algorithm over factorial speech processing models (this can be considered as an extension to the single chain token passing algorithm by using the two dimensional Viterbi algorithm for performing joint-decoding in factorial speech processing models).**

Token scores are mainly affected by the joint-state likelihoods rather than state transition probabilities in each acoustic model. Moreover since in the challenge appearance of words in each word slot are equally likely, no other score updating due to this is involved in the steps of algorithm which can be considered for more general tasks.

The algorithm can support multiple phonetic transcription of a word. Also we can consider multiple chains in the factorial models of speech processing. The extension is straightforward by considering that going forward in the time must be performed chain-by-chain and then returning to the first chain; similar to the two-dimensional Viterbi algorithm. In fact, in the two-dimensional Viterbi algorithm, the dynamic programming is run within a time-frame in addition to running in time, which reduces the computational complexity of inference.

## 3. Determining and adapting source models

Using speaker adapted source models in the factorial models for the mixed-speech recognition task of "monaural speech separation and recognition challenge", significantly improves recognition performance. This happens since discriminative features of speaker voices are also clues for improving acoustic inference within time-slices rather than dynamic constraints which applies to the whole utterance. Additionally, any adaptation related to the gain effect of the expression (1) for synthetizing the mixed-speech signal also significantly improves recognition performance. In the mixed-speech recognition task of the challenge, utterances of two speakers among 34 speakers of the GRID dataset are randomly selected for synthetizing each mixed-speech signal. During the test phase, using speaker labels of the test files is prohibited and therefore the identity of speakers for each utterance is unknown to the recognizer. On the other hand, TMR is also unknown to the recognizer. Identifying speakers and gain estimation is needed for determining audio source models of factorial models and adapting model parameters for improving inference and recognition performance which are described in the next two sub-sections.

### 3.1. Speaker identification

In the challenge, model based methods such as the IBM system use speaker conditional likelihoods of each frame for the calculation of speaker id posteriors over the frames of the input mixed-speech signal. Then the speaker identification is done by voting high confidence identified frames which yields relatively precise speaker identification. This method is applied based on an underlying assumption which states that for the high resolution spectral features we can assume that one source is dominant in each time-frequency cell after the short-term discrete Fourier transform analysis. In the proposed method, we simply use a deep neural network for joint-speaker identification which also yields competitive results with relative simplicity and lower test-time computations compared to the previous methods.

A deep feed forward architecture with 34 sigmoid output neurons are selected for construction of the network. In addition, high resolution Mel-scale filterbank energies in a context-window including $2\tau + 1$ frames are selected as the network input (see Fig. 7). High resolution features of the synthetized mixed-speech signals with speaker identities are provided for the training phase. For each mixed-speech signal, two output neurons are designated as the desired output value except for the case where one speaker is presented as both target and masker in the signal. The network architecture and its input and target values are illustrated in Fig. 7. All neuron activation functions have logistic sigmoid shape.

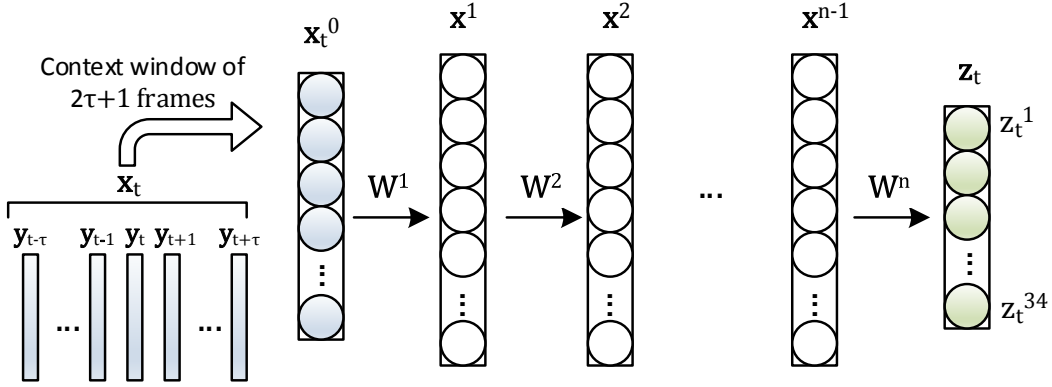

**Fig. 7. Joint-speaker identification network.**

The joint-speaker identification network is pre-trained layer-wise by restricted Boltzmann machines (RBM) generatively, which is necessary before the training phase. Then the network is trained by error back-propagation using speaker labels.

In the test-phase, for each context-window of frames of each utterance, network output is extracted. At first, uncertain neurons (neurons with low activation values) are eliminated by a threshold as follows:

$$z_t^i = \begin{cases} 0 & z_t^i < \lambda \\ z_t^i & z_t^i \geq \lambda \end{cases}, \forall\, i \in \{1,2,3,\dots,34\}, t \in \{1,2,3,\dots,T\} \qquad (24)$$

Then for these context-windows, two best speaker ids are selected for final voting as follows:

$$s_t^j = \begin{cases} z_t^j & j \in \{c_1, c_2\} \\ 0 & ow \end{cases} \qquad (25)$$

in which $c_i$s are the indices of sorted activation values within a context-window; i.e. **sort**$((z_t^i)_{i=1}^{34}) = (z_t^{c_i})_{i=1}^{34}$. Now speaker voting is done by the following summation:

$$s^i = \sum_{t=1}^{T} s_t^i \qquad (26)$$

At this step speaker scores are sorted and speaker indices are stored in $c_i$s; i.e. **sort**$((s^i)_{i=1}^{34}) = (s^{c_i})_{i=1}^{34}$. Then, recognized speakers are selected using a second threshold, $\theta$, and the limit of maximum two speakers. The $\theta$ threshold is used to eliminate the second speaker in the same speaker test utterance cases.

$$\{c_i | s^{c_i} > \theta, \forall i \in \{1,2\}\} \tag{27}$$

Two thresholds, $\lambda$ and $\theta$ can be tuned by the validation set which will be discussed in the experiments.

### 3.2. Gain estimation and model adaptation

Recalling the expression (1) for the generation of mixed-speech signal, relative speech signal energy of the speaker $a$ and $b$ can be modeled by a gain coefficient ($g$). Here, at first, by considering one audio source, the effect of gain coefficient in each step of feature extraction is investigated. Starting from the power spectrum of speech frames ($|x|^2$), for extracting MFCC features we have:

$$^{c}x' = C \log(W|gx|^2) \tag{28}$$

again, $g$ is the gain, $W$ is the matrix of Mel shaped averaging filters, and $C$ is the [truncated] DCT matrix. Then we have:

$$^{c}x' = C(\log(W|x|^2) + 2\log(g\mathbf{1})) = {}^{c}x + x_0 \tag{29}$$

where $\mathbf{1}$ is column vector of one and $^{c}x$ is the original MFCC features. Then, $x_0$ is the gain compensated vector as:

$$x_0 = 2\log(g)\, C \times \mathbf{1} = [g_0, 0, 0, \ldots, 0]^T \tag{30}$$

where the $g_0$ is defined as:

$$g_0 = 2\log(g)\sqrt{m} \tag{31}$$

in which $m$ is the number of Mel filters in the filterbank (the number of columns of $W$).

As we can see in (30), gain factor only affects the first MFCC coefficient (MFCC's 0th order coefficient). For the delta and acceleration features based on (13), we can observe that the gain has no effect on delta coefficients and consequently on acceleration coefficients. Therefore, only the first element of each feature vector is affected by the gain coefficient and we must adjust the related parameters of the source models for the adaptation. Equation (29) shows that a constant vector is added to the original MFCC features. Therefore, only Gaussian mean vectors must be updated and covariance matrices remain unaltered. In fact, only the first element of Gaussian means in the source models must be adjusted.

In the test phase, the gain coefficient and therefore $g_0$ is unknown and it must be estimated. The general method for this estimation is to use maximum likelihood estimators where for the appropriate gain coefficient, the likelihood of test utterance when decoded in the adapted model is maximized. i.e.:

$$\hat{g}^* = \underset{g_i}{\operatorname{argmax}} \mathcal{L}^*(\mathbf{y}_{1:T}|g_i) \tag{32}$$

The method is applicable in this way, because factorial speech processing models are generative models which model the way clean audio sources are mixed for creating output signals. Therefore, a more matched parameter set yields to a greater decoding likelihood for the test utterance. During the challenge, model based methods have used this technique in their gain estimation step.

In our work, we use deep architectures for this step once again. In fact, similar architecture to joint-speaker identification is used here for gain estimation except that a linear neuron is used for estimating the gain coefficient. During the training phase, mixed-speech training utterances with various gains are synthetized for the network with its gain as the desired output. Feature extraction and framing is similar to joint-speaker identification by deep networks. This method for estimating the masker gain is straightforward compared to the past MLE methods. However, it is applicable to the task of this paper, since only the gain of the masker utterance is adjusted in this task. At the final step, based on the estimated gain, model parameters are updated and inference is done for decoding.

## 4. Experiments

The monaural speech separation and recognition challenge is designed based on the GRID dataset (Cooke et al., 2006). Speech material of this task consists of 1000 utterances of 34 speakers which are partitioned evenly for the training and test phase. Synthetized mixed-speech of randomly selected joint-speakers are extracted from 500 utterances of the test part for each speaker in 6 different TMRs including 6, 3, 0, -3, -6 and -9 dBs. For each TMR, 600 and 300 mixed-speech signals are synthetized for the test and development set respectively. While identification of joint-speakers and the mixed-speech TMR can be understood by the filenames and folders, it cannot be used in the test phase. In this task, recognition of the letter and digit of the target speaker is the objective and the scoring is done by the scoring scripts which are provided by the organizers. The target speaker always uses the "white" color and the masker does not and we can use this clue to identify the target speaker. For this challenge, recognition performance of human listeners is evaluated for comparing automated systems with humans as the gold standard (Fig. 11).

In this section, source modeling, feature extraction, grammar definition and extraction of the task dictionary are explained. Then based on experiments conducted on the development set, selection of feature space and tuning speaker identification hyperparameters are done and the results are provided. Finally, the main experiments are conducted and the results are compared to the past super-human results.

### 4.1. Source Models, Grammar and Lexicon

Three state HMM monophones are selected as source acoustic models which are trained by the HTK tool (Young et al., 2009). The models are first initialized by the TIMIT dataset (Garofolo et al., 1993), then re-estimated by training speech utterances from 34 speakers of the challenge corpus. The number of mixture components is increased from 8 to 32 for each monophone state. Then models are adapted for each speaker and the number of mixture components is decreased to 8 and 4 components which are used in the experiments. The four component speaker adapted models are used for feature selection and phase factor adjustments and the eight component models are used for the main experiments. MFCC features are used for acoustic modeling and feature extraction is done by the Voicebox toolbox (Brookes, 1997). For feature extraction, 27 Mal-scale filters are used and framing is done by 10ms frame shift and 25ms frame length. Number of 13 through 23 MFCC coefficients (excluding delta and acceleration coefficients) are extracted using the truncated DCT matrix (its effect will be discussed further in the experiments).

HMM initialization by the TIMIT dataset requires no Grammar and Dictionary and 40 monophones (including silence model) are trained by the TIMIT dataset. For HMM re-estimation, due to the word level transcription of training speech materials, the definition of lexicon and Grammar is required. We use the BEEP dictionary (Robinson, 1997) for the phonetic transcription of 51 different words of the challenge (see Fig. 1). Additionally, the word lattice network is created based on the task grammar which is already provided in (Cooke et al., 2010) and also presented in Fig. 1.

## 4.2. Feature selection and phase factor determination

Before initiating the main experiments, selection of appropriate features for this task and adjusting the phase factor in the mismatch function of (9) for performing the acoustic inference must be done. At this step, MFCC features and their first and second order derivatives are selected as the feature type. Now, various static phase factors are selected for the test on the development set. Configuration of feature extraction is similar to the task baseline system except that MFCC 0th's coefficient is used in features instead of the logarithm of frame energy and 27 filters are used in the Mel-scale filter bank.

At this step, no speaker identification and gain adaptation is carried out and identity of the speakers are considered to be known during the tests. Since at this step, gain adaptation is not performed yet, only performance of the system in near zero TMRs are considered for phase factor (alpha) selection; i.e. 3, 0 and -3 dB TMRs. Fig. 8 shows the average performance of these three TMRs against different alpha values. The best alpha value is selected for the next experiments. Alpha candidates are selected almost similar to the work of (Li et al., 2009) and the selected alpha is 2. This alpha value is not valid regarding its support set (the reader is referred to (Li et al., 2009) or (Van Dalen, 2011) for explanations of this theoretical contradiction).

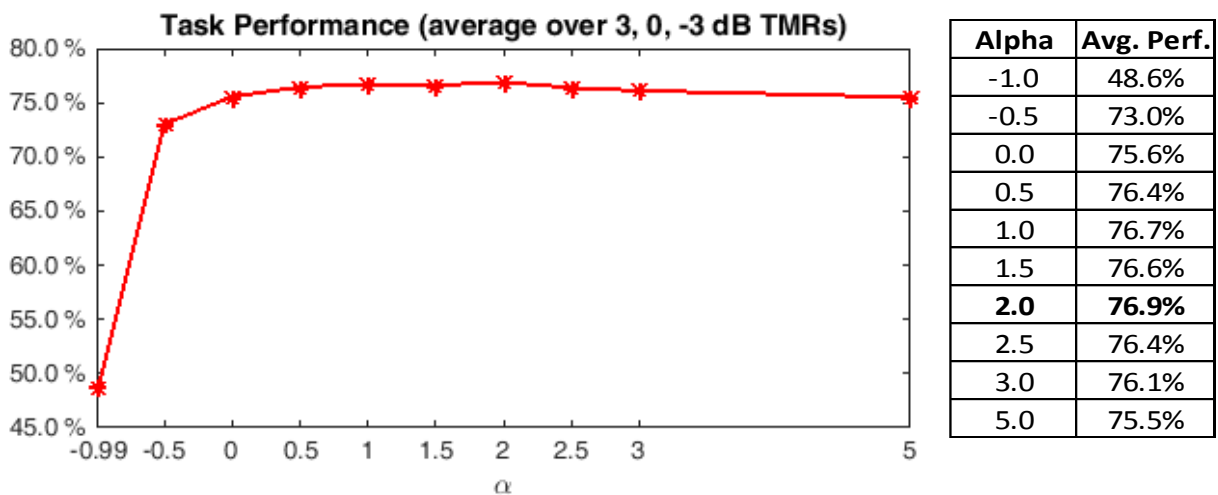

| Alpha | Avg. Perf. |
|---|---|
| -1.0 | 48.6% |
| -0.5 | 73.0% |
| 0.0 | 75.6% |
| 0.5 | 76.4% |
| 1.0 | 76.7% |
| 1.5 | 76.6% |
| **2.0** | **76.9%** |
| 2.5 | 76.4% |
| 3.0 | 76.1% |
| 5.0 | 75.5% |

**Fig. 8. The average task performance over TMRs 3, 0 and -3 dB against different static phase factors (alpha values).**

For selecting the appropriate feature type, a different combination of MFCC features are selected for evaluation on the development set which their results are listed in Table 1. These feature types are used for source modeling and acoustic inference with the selected alpha value. Again here, only system performance is considered for near zero TMRs.

Table 1: List of features with their properties for evaluation on the development set for feature type selection. Average performance over 3, 0 and -3 dB TMRs are listed in the last column.

| Feature type | Truncated DCT Matrix Size | Including delta coeff. | Including accel. coeff. | Total feature dimension | Clean speech recognition performance | Average Task Performance |
|---|---|---|---|---|---|---|
| **MFCC0(13)** | 13x27 | no | no | 13 | 95.16 % | 74.78 % |
| **MFCC0(26)** | 26x27 | no | no | 26 | 96.46 % | 75.22 % |
| **MFCC0D(26)** | 13x27 | yes | no | 26 | 97.75 % | 77.00 % |
| *MFCC0D(38)* | 19x27 | yes | no | 38 | 98.01 % | **80.16 %** |
| **MFCC0D(42)** | 21x27 | yes | no | 42 | 98.08 % | 78.95 % |
| **MFCC0D(46)** | 23x27 | yes | no | 46 | 97.72 % | 79.56 % |
| *MFCC0DA(39)* | 13x27 | yes | yes | 39 | 97.76 % | 76.89 % |

Two feature types are selected for the main experiments. The first is the de facto MFCC0DA(39) and the second is MFCC0D(38).

### 4.3. Speaker identification results

Joint-speaker identification of this task is done by a deep neural network. The network contains 5 layers, including 4 hidden layers (each layer has 2500 logistic sigmoid neurons) and one output layer which consists of 34 neurons for joint-speaker identification. The network is fed with concatenation of 21 high resolution power spectral features ($\tau = 10$) which are extracted by installing a dense Mel-scale filterbank on the frame power spectrum (110 filters are included in the filterbank). Pre-training of the network is done layer-wise by RBMs and then it is fined-tuned with the speaker labels by the DeeBNet toolbox (Keyvanrad and Homayounpour, 2014). About 7000 training utterances are synthetized by mixing target and masker speaker utterances extracted from the training data in different TMR values similar to the test data. The related speaker labels are attached to the training data for the network fine-tuning.

Speaker adaptation hyperparameters , $\lambda = 0.99$ and $\theta = 0.1$, are adjusted by the development set according to (27). Based on the selected thresholds, speaker identification results are summarized in Table 2 and compared to the IBM system speaker identification results.

Table 2: Joint-speaker identification accuracy of the proposed method compared to the past IBM super-human system in different TMRs. It can be seen that while the proposed method performs the job in one phase it has competitive results to the past two-phase system.

| TMR (dB) | -9 | -6 | -3 | 0 | 3 | 6 | Avg. |
|---|---|---|---|---|---|---|---|
| **The Proposed System (DNN based, one-phase)** | 94.3 % | 98.0 % | 99.3 % | 99.3 % | 99.0 % | 95.0 % | 97.5 % |
| **IBM System (GMM based, two-phase)** | 96.5 % | 98.1 % | 98.2 % | 99.0 % | 99.1 % | 98.4 % | 98.2 % |

The IBM system, uses GMM speaker models and speaker likelihoods for speaker identification in two steps. At first, a list of speaker candidates (called finalists) are selected based on speaker id posteriors. In the second phase, different combinations of joint-speakers are selected from the speaker finalists to then performing a search for finding the best joint-speakers beside their associated gain. Joint-speaker ids which their gain adapted models maximize utterance likelihood are selected as the identified speakers. Comparing our system to IBM's super-human system, it can be seen that competitive results are achieved in a single phase identification. As another observation, we can see that in near zero TMRs both systems perform better than the cases when one speaker dominates the other (speech of one speaker masks the other speaker and prevents correct identification).

**4.4. Joint-decoding by gain adapted models, pushing forward past super-human results**

For the main experiments, speaker identification and TMR information cannot be used during the decoding. Joint-identification of speakers and gain estimation is done by the trained deep neural networks and the rest of the recognition procedure is carried as mentioned before which is illustrated in Fig. 9.

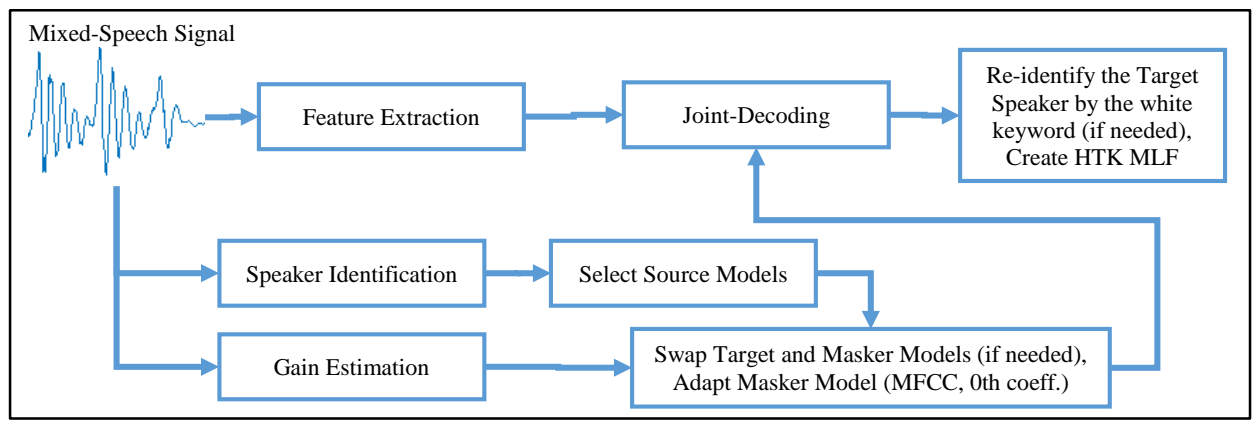

**Fig. 9. Recognition procedure for mixed-speech signals.**

The role of the two target identifiers shown in this figure are explained as follows. The joint-speaker identification network can identify those speakers which have some speech footprints in the mixed-speech signal, even in extreme TMRs; i.e. 6, -6 and -9 dBs. But it cannot determine the masker speaker which is needed for further model adaptation. For resolving this ambiguity, the gain adapted model which maximizes the likelihood of the test utterance is selected as the masker model. After that, joint-decoding is performed based on the proposed algorithm (see Fig. 6). At this step, another check for the possible target-masker swapping is done for reporting the recognized letter and digit of the target speaker. This step provides no change on extreme TMRs, since the adapted masker model can distinguish between the target and masker utterances during the joint-decoding.

Recognition results of the selected feature types based on the proposed procedure are provided in Table 3. Additionally, recognition results when we select and adapt source models based on test file information (oracle speaker ids) are also provided in this table which removes the effect of the speaker identification phase. As it was expected based on the results of Table 1, the MFCC-delta features perform better than the MFCC-delta-acceleration features. This can be explained by the approximation involved during the acoustic inference of the dynamic coefficients. The approximation assumes that source states are not changed during the extraction of dynamic coefficients of each frame. Moreover, by comparing the results to the oracle system, we can see more performance upgrade for this task is possible when we use a more accurate speaker identifier.

**Table 3: Recognition results of the two selected feature types. It can be seen that acceleration coefficients gain no performance improvement over MFCC-delta features. Moreover recognition results when we use oracle speaker identifier are provided in this table.**

| TMR (dB) Feature Type | -9 | -6 | -3 | 0 | 3 | 6 | Avg. |
|---|---|---|---|---|---|---|---|
| **MFCC0D(38), oracle speaker ids** | 79.8 % | 85.0 % | 84.8 % | 83.4 % | 89.5 % | 92.5 % | 85.8 % |
| **MFCC0D(38)** | **76.6 %** | **83.2 %** | **83.3 %** | **81.9 %** | **88.4 %** | **89.9 %** | **83.9 %** |
| **MFCC0DA(39)** | 70.6 % | 78.3 % | 80.0 % | 79.6 % | 85.9 % | 86.6 % | 80.2 % |

Detailed recognition accuracy for MFCC-delta features for different TMRs over three sets of test utterances is also provided in Table 4. As it can be seen, speaker difference clues can improve recognition accuracy which is apparent when the results of the "Same Talker" condition are compared to the other cases. Additionally, on the other axis of the "Same Talker" condition we can see that the only distinguishing factor is the relative energy of voices of target and masker speakers. In this case, recognition performance is degraded in near zero conditions (see Fig. 10 for the gain adapted case). On the other hand, it was expected that the best results can be achieved in "Different Gender" test utterances for the best TMRs, i.e. 3 and 6 dBs.

**Table 4: Detailed recognition accuracy for MFCC-delta features (MFCC0D-38) for different TMRs over three sets of test utterances.**

| TMR (dB) Joint-Speaker Sets | -9 | -6 | -3 | 0 | 3 | 6 | Avg. |
|---|---|---|---|---|---|---|---|
| **Same Talker** | 74.7 % | 77.2 % | 72.4 % | 64.5 % | 79.2 % | 84.6 % | 75.4 % |
| **Same Gender** | 79.3 % | 88.3 % | 89.1 % | 91.6 % | 92.7 % | 92.2 % | 88.9 % |
| **Different Gender** | 76.3 % | 85.3 % | 90.3 % | 92.5 % | 94.8 % | 93.8 % | 88.8 % |
| **Overall** | 76.6 % | 83.2 % | 83.3 % | 81.9 % | 88.4 % | 89.9 % | 83.9 % |

A further comparison for decoding results of gain adapted source (masker) models and unaltered models can be done by investigating Fig. 10. It can be seen that gain adaptation is substantially beneficial in the same talker condition. This is due to the fact that both source models are the same and no other discriminative clue is available in this case except their first dimension parameters which is changed in the gain adaptation phase. Moreover, it can be seen that gain adaptation is more beneficial in extreme TMR conditions in all test sets and has no benefit in zero TMR. In the zero TMR case, some minor adaptation may occur due to the variation of the volume of speaker voice in different recording trials.

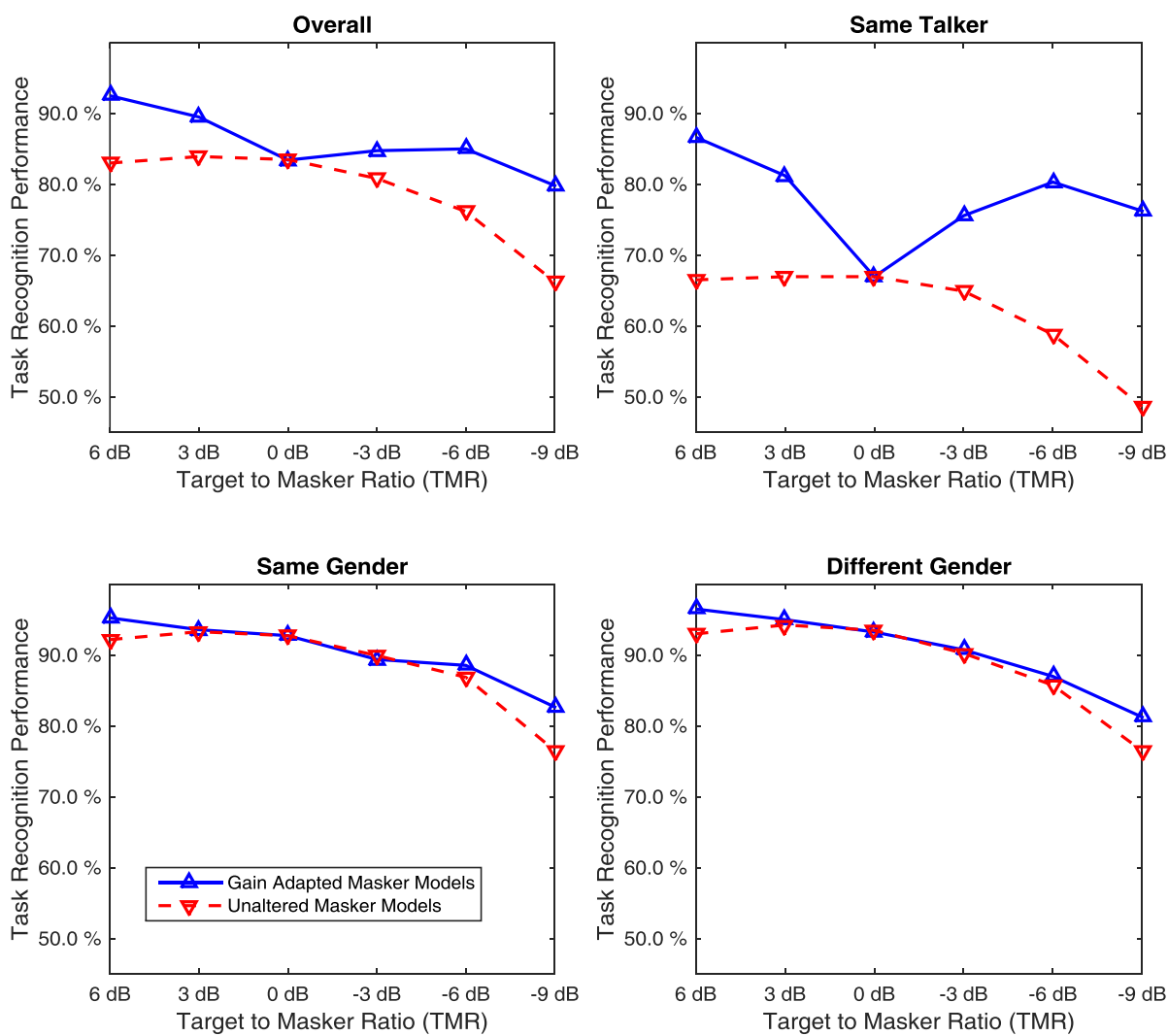

Fig. 10. Effect of masker model gain adaptation by comparing gain adapted model results and gain ulaltered masker model results in different test sets. It can be seen that gain adaptation is substantially beneficial in the same talker condition. Even in this case in zero TMR, as we expected, no improvement is achieved by some minor gain adaptations. In these tests, oracle speaker ids are used for source model selection.

For the final comparison, the past best-achieved results for this challenge, to the best of our knowledge, are provided in Fig. 11. The first super-human system was the IBM system (Hershey et al., 2010; Rennie et al., 2010) and the second one was the recent Microsoft DNN based recognizer (Weng et al., 2015). The IBM super-human system has two phases for performing recognition. The first phase is to separate speech of the target and masker speakers, which its results were amazing at that time, especially their extension to the task which supports up to five interfering speakers. Then based on the separated speech it performs decoding. The speech separation phase of the IBM system was done based on factorial models. In their system, the feature vector state-conditional posteriors are calculated in the inference phase for the speech reconstruction. In their work, various configurations are considered for this task, but their best performance is provided in Table 5 and also plotted in Fig. 11.

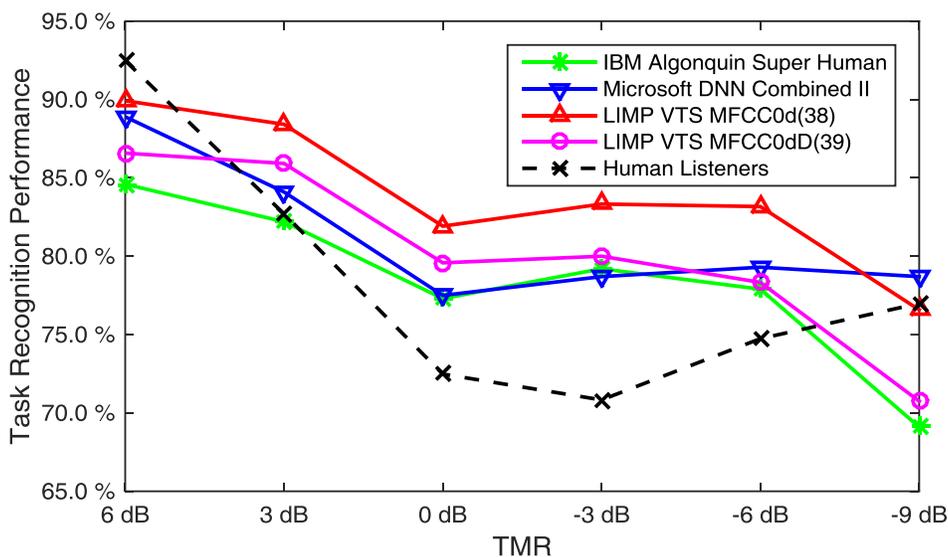

**Fig. 11. Comparison of results of our implemented system (LIMP VTS based system) and the past super-human results. Human listener performance is also provided in this figure. The plot shows that in all TMRs, except -9 dB, the proposed method outperforms the two other systems. For the extreme TMR of -9 dB, the energy difference between the sources makes the Microsoft DNN based system superior to the proposed system. This is due to using the difference of the frame's instantaneous energy which is significant in this TMR. Moreover, it can be seen that the DNN based system performs more similar to human listeners in extreme TMR conditions (6, -6 and -9 dB).**

Recently, a DNN based recognizer from Microsoft research outperforms the IBM system. The system was not based on speaker adapted models but it uses energy difference clues by training a set of two DNNs for senone posterior extraction. One DNN is trained for high energy frames and the other for low energy frames. It can be seen that this system performs well in extreme TMRs where it can incorporate more energy difference clues in the mixed-speech signal. Since it uses instantaneous frame energy and it controls energy switching time-stamps, near zero TMR results are also surprising. Comparing the results to human listeners, it can be seen that human listeners cannot perform well in near zero TMRs while highly specialized joint-decoding systems can perform far better than humans in these conditions.

Table 5: Recognition results of IBM and Microsoft super-human systems. The last row is the result of the proposed system (also provided in Table 3 and in details in Table 4).

| TMR (dB)<br>Feature Type | -9 | -6 | -3 | 0 | 3 | 6 | Avg. |
|---|---|---|---|---|---|---|---|
| **IBM factorial based system** | 84.6 % | 82.2 % | 77.3 % | 79.2 % | 77.9 % | 69.1 % | 78.4 % |
| **Microsoft DNN based system** | 88.9 % | 84.1 % | 77.5 % | 78.7 % | 79.3 % | 78.7 % | 81.2 % |
| **LIMP factorial based system** | 76.6 % | 83.2 % | 83.3 % | 81.9 % | 88.4 % | 89.9 % | 83.9 % |

## 5. Conclusion

In this paper, factorial speech processing models were presented by the language of probabilistic graphical models. Conditional probability distributions of these models are described; especially detailed derivation of their centric CPD, the CPD which combines source audio features. Moreover, the inference algorithm over these models is derived using the factor graphs. The idea of token passing for large vocabulary continuous speech recognition is extended in this work to support decoding of the task of monaural speech separation and recognition challenge. Therefore, a joint-token passing algorithm based on the idea of token passing is developed to perform joint-decoding on factorial speech processing models by the two-dimensional Viterbi algorithm. Moreover, a set of two task specific deep neural networks is suggested and used in this work for joint-speaker identification and gain estimation. Based on these networks, source speaker models are first selected and then adapted for compensating gain mismatch for the masker speaker in the test phase of the challenge. We can see that the developed single phase joint-speaker identification network makes the joint-speaker identification comparable to the past best system in the challenge. Additionally, we can see that gain adaptation is very effective in improving the system performance simply by adjusting one dimension of masker source models.

The proposed system of this work outperforms two past super-human systems for this challenge. Comparing it to the IBM factorial system, this superiority is due to performing direct inference over the mixed-speech signal which incorporates all uncertainty during the inference. Moreover, source modeling of low resolution feature spaces in our work is more accurate and convenient rather than facing with high resolution spectral features which was necessitated by the separation phase of the IBM system. Comparing it to the Microsoft DNN based recognizer, since we used speaker adapted models rather than only high and low energy DNNs, the proposed system performs better than the DNNs by using discriminating information of the speakers. In a fair comparison, it can be mentioned that the Microsoft system uses only two main DNN models which is a more generic solution than our task specific speaker adapted models. As a future work, it is expected that by performing DNN based acoustic inference using speaker adapted DNNs, even the best-achieved results of this task can be more improved. The problem is to develop a mechanism for training jointly speaker adapted DNNs for performing acoustic inference.

## Acknowledgements

The authors would like to thank Mr. Mohammad Ali Keyvanrad for his valuable arguments and support of DeeBNet toolbox in this work.